\documentclass[11pt]{article}

\usepackage[preprint]{neurips_2023}

\usepackage{times}
\usepackage{latexsym}
\usepackage{makecell}
\usepackage[T1]{fontenc}

\usepackage[utf8]{inputenc}

\usepackage{microtype}
\usepackage{breakurl}
\usepackage{times}
\usepackage{latexsym}

\usepackage{microtype}
\usepackage{hyperref}

\usepackage{graphicx}
\usepackage{xspace}
\usepackage{paralist}
\usepackage{booktabs}
\usepackage{multirow}
\usepackage{amsmath}
\usepackage{amsfonts}
\usepackage{amssymb}
\usepackage{pifont}
\usepackage{mathrsfs}
\usepackage{algorithm,algpseudocode}
\usepackage{mdwlist}
\usepackage{enumitem}
\usepackage{color}
\usepackage{dsfont}
\usepackage{amsthm}
\usepackage{bm}
\usepackage{float}
\usepackage{caption}
\usepackage{subcaption}
\usepackage{hyperref}
\usepackage{tikz}
\usepackage{tcolorbox}
\usepackage{caption}
\usepackage{multicol}
\newcommand{\method}{SpeechGPT\xspace}

\DeclareCaptionJustification{mycenter}{\centering}

\title{\Large\method: Empowering Large Language Models with Intrinsic Cross-Modal Conversational Abilities }

\author{
    \textbf{Dong Zhang},
    ~\textbf{Shimin Li},
    ~\textbf{Xin Zhang},
    ~\textbf{Jun Zhan},
    ~\textbf{Pengyu Wang},\\
    ~\textbf{Yaqian Zhou}\thanks{Corresponding author},
    ~\textbf{Xipeng Qiu}\footnotemark[\value{footnote}] \\
    \\
    School of Computer Science, Fudan University\\
    Shanghai Key Laboratory of Intelligent Information Processing, Fudan University\\
    {\tt 	dongzhang22@m.fudan.edu.cn} \\
    {\tt 	\{smli20,zhouyaqian,xpqiu\}@fudan.edu.cn} \\
    \\\\
    \url{https://github.com/0nutation/SpeechGPT}
}

\begin{document}
\maketitle

\begin{abstract}

Multi-modal large language models are regarded as a crucial step towards Artificial General Intelligence~(AGI) and have garnered significant interest with the emergence of ChatGPT. However, current speech-language models typically adopt the cascade paradigm, preventing inter-modal knowledge transfer. In this paper, we propose \method, a 
large language model with
intrinsic cross-modal conversational abilities, capable of perceiving and generating multi-model content. With discrete speech representations, we first construct SpeechInstruct, a large-scale cross-modal speech instruction dataset. Additionally, we employ a three-stage training strategy that includes modality-adaptation pre-training, cross-modal instruction fine-tuning, and chain-of-modality instruction fine-tuning. The experimental results demonstrate that SpeechGPT has an impressive capacity to follow multi-modal human instructions and highlight the potential of handling multiple modalities with one model. Demos are shown in \url{https://0nutation.github.io/SpeechGPT.github.io/}.
\end{abstract}

\section{Introduction}
Large language models ~\citep{openai2023gpt4, touvron2023llama} have performed astonishingly on various natural language processing tasks. Meanwhile, multi-modal large language models, such as GPT-4, PALM-E~\citep{driess2023palm}, and LLaVA~\citep{liu2023visual}, have explored the ability of LLMs to understand multi-modal information. 
However, a significant gap exists between current LLMs and general artificial intelligence (AGI). First, most current LLMs can only perceive and understand multi-modal content but cannot spontaneously generate multi-modal content. Second, continuous signals like images and speech cannot be adapted directly to LLMs that receive discrete tokens.

\begin{figure}[t]
\centering
\includegraphics[width=0.8\columnwidth]{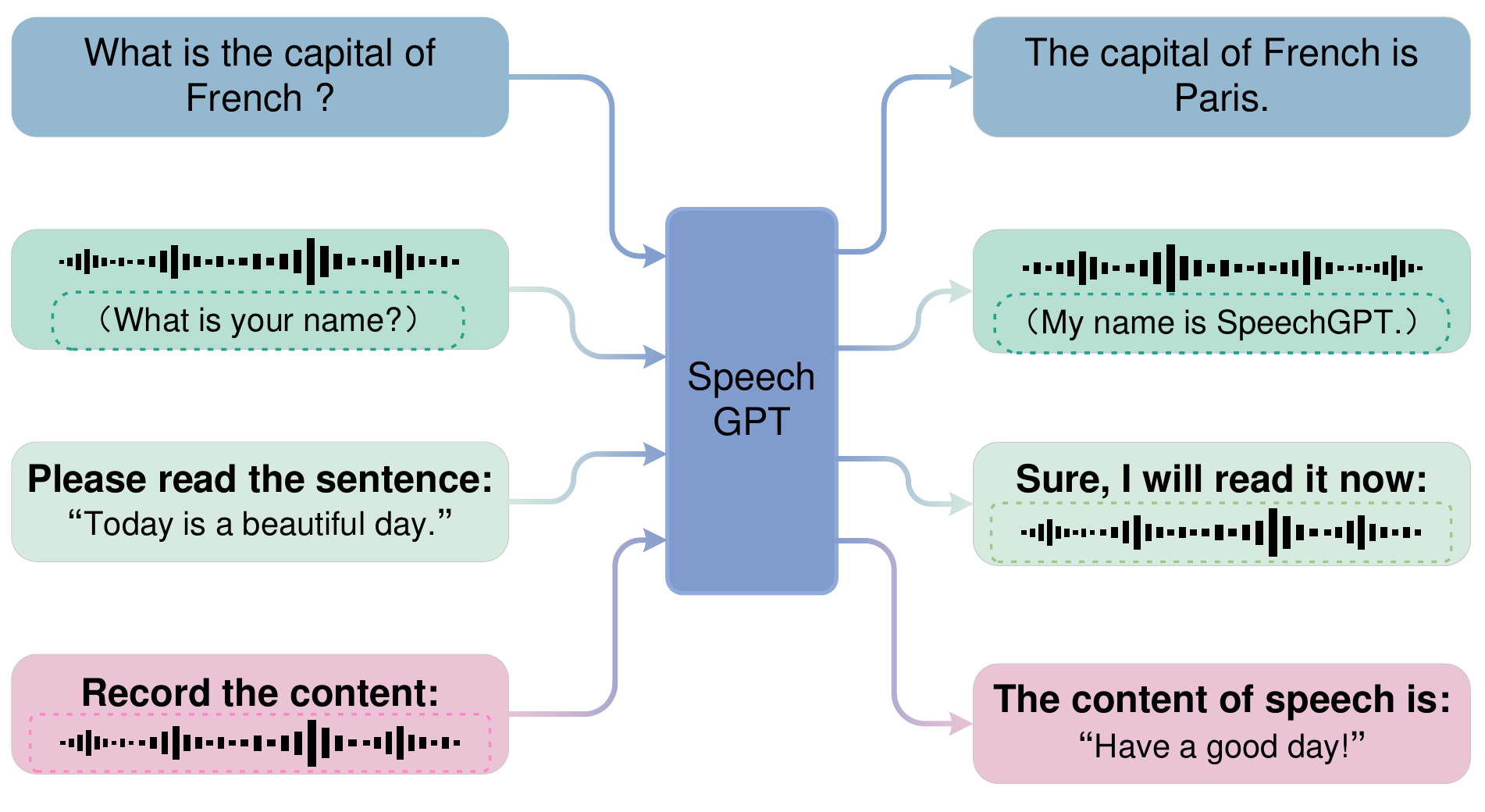} 
\caption{SpeechGPT's capabilities to tackle multiple cross-modal tasks.}
\label{fig:task_intro}
\end{figure}

The current speech-language model mainly adopts a cascading paradigm~\citep{huang2023audiogpt}~\, i.e., the LLM is connected with an automatic speech recognition (ASR) model or a text-to-speech (TTS) model in tandem, or the LLM is employed as a control hub, with several speech processing models are integrated to cover multiple audio or speech tasks~\citep{huang2023audiogpt, shen2023hugginggpt}. Some prior work on generative spoken language models involves encoding the speech signal into a discrete representation~\citep{baevski2020wav2vec, hsu2021hubert} and modeling it with language models~\citep{lakhotia2021generative,borsos2022audiolm, zhang2023speak, wang2023neural}.

While capable of perceiving and generating speech, the existing cascading methods or spoken language models still have several limitations. First, the LLM in the cascaded model only functions as a content generator. Since the representations of speech and text are not aligned, the LLM's knowledge cannot be transferred to the speech modality. Second, the cascade approach~\citep{shen2023hugginggpt, huang2023audiogpt} suffers from the loss of paralinguistic signals such as emotion and prosody. Third, existing spoken language models~\citep{wang2023neural, zhang2023speak} only synthesize speech but fail to comprehend its semantic information, preventing them from achieving true cross-modal perception and generation.


In this paper, we propose \method, a large language model with
intrinsic cross-modal conversational abilities, capable of perceiving and generating multi-model content.
We perform speech discretization with a self-supervised trained speech model to unify the modality between speech and text. The discrete speech tokens are then expanded into the vocabulary of the LLM, thus endowing the model with an inherent competence to perceive and generate the speech.

To provide the model with the capacity to handle multi-modal instructions, we build the first speech-text cross-modal instruction-following dataset SpeechInstruct. Specifically, we discretize the speech to discrete units~\citep{hsu2021hubert} and construct the cross-modal unit-text pair based on the existing ASR dataset. Meanwhile, we construct hundreds of instructions for diverse tasks with GPT-4 to simulate actual user instructions as illustrated in Appendix~\ref{sec:app:task_desc}. In addition, to further enhance the model's cross-modal capability, we designed the Chain-of-Modality instruction data, i.e., the model receives the speech command, thinks about the process in text, and then outputs the response in speech.

For better cross-modal transfer and efficient training, SpeechGPT undergoes a three-stage training process: modality-adaptation pre-training, cross-modal instruction fine-tuning, and chain-of-modality instruction fine-tuning. The first stage enables speech comprehension for SpeechGPT with the discrete speech unit continuation task. The second stage employs the SpeechInstruct to improve the model's cross-modal capabilities. The third stage utilizes parameter-efficient LoRA~\citep{hu2021lora} fine-tuning for further modality alignment.

To evaluate the effectiveness of SpeechGPT, we conduct a wide range of human evaluations and case analyses to estimate the performance of SpeechGPT on textual tasks, speech-text cross-modal tasks, and spoken dialogue tasks. The results demonstrate that SpeechGPT exhibits a strong ability for unimodal and cross-modal instruction following tasks as well as spoken dialogue tasks.


Our contributions include the following:
 \begin{itemize}[itemsep=1pt, leftmargin=10pt, parsep=0pt, topsep=1pt]
    \item 
    We build the first multi-modal large language model that can perceive and generate multi-modal contents.

    \item 
    We construct and release SpeechInstruct, the first large-scale speech-text cross-modal instruction-following dataset.

    \item 
    We build the first spoken dialogue LLM with strong human instruction following ability and spoken dialogue ability.

    \item 
    We show great potential to incorporate other modalities into LLMs through discrete representations.

\end{itemize}




\section{Related Work}

\noindent\textbf{Multi-modal Large Language Model}~
 Current multi-modal LLMs predominantly focus on the visual domain, feeding continuous representations obtained from pre-trained visual encoders into LLMs, facilitating full-parameter or parameter-efficient training on visual-language data~\citep{openai2023gpt4, huang2023language, zhang2023llamaadapter}.
Palm-E~\citep{driess2023palm} integrates the 540B PaLM~\citep{chowdhery2022palm} and 22B Vision Transformer~\citep{dosovitskiy2021image} into the largest vision-language model.
LLaVA~\citep{liu2023visual} leverages pre-trained CLIP~\citep{radford2021learning} visual encoder and LLaMA~\citep{touvron2023llama} and conduct instruct tuning on GPT4-assisted visual instruction data.
X-LLM~\citep{chen2023xllm} converts multi-modalities into representations with X2L interfaces as the inputs of the large language model.
However, such structures only enable LLMs to process multi-modal input, without ability to generate multi-modal output. Diverging from prior studies, our approach emphasizes the development of a speech-centric multi-modal LLM, endowing it with the proficiency to accommodate both multi-modal input and output.

\noindent\textbf{Generative Spoken  Language Model}~
Discrete self-supervised representation based spoken generative language modeling is making remarkable progress on large-scale speech dataset training~\citep{nguyen2022generative}.
AudioLM~\citep{borsos2022audiolm} proposes to model speech based on audio codecs together with semantic codes, which can synthesize speech in a textlesss setting. VALL-E~\citep{wang2023neural} builds a generative spoken language model on audio codecs and treat Text-to-Speech as a conditional generation task. However, these models are designed for a specific task and failed to benefit from LLMs. \method is built upon the foundation of LLM and transfers LLM’s knowledge to speech modality, consequently obtaining better task generalization and human-instruction following ability.

\noindent\textbf{Speech-Enabled LLM Interaction}~
Following the emergence of ChatGPT, several studies have concentrated on the integration of expert speech models with LLMs to enable direct speech interaction with LLMs. HuggingGPT~\citep{shen2023hugginggpt} facilitates task decomposition of human instructions by LLMs and allows the invocation of models from Huggingface to accomplish specific tasks, encompassing a range of automatic speech recognition (ASR) and text-to-speech models. AudioGPT~\citep{huang2023audiogpt} leverages a variety of audio foundation models to process complex audio information and connect LLMs with input/output interface
(ASR, TTS) for speech conversations. However, these models exhibit increased complexity, demand extensive resources, and are prone to the unavoidable error accumulation problems. Our approach enables speech interaction with LLMs without relying on ASR or TTS systems, circumventing the aforementioned drawbacks.



\begin{figure*}[t] 
    \setlength{\abovecaptionskip}{-0.cm}
    \setlength{\belowcaptionskip}{-0.5cm}
    \centering 
    \includegraphics[width=1\textwidth]{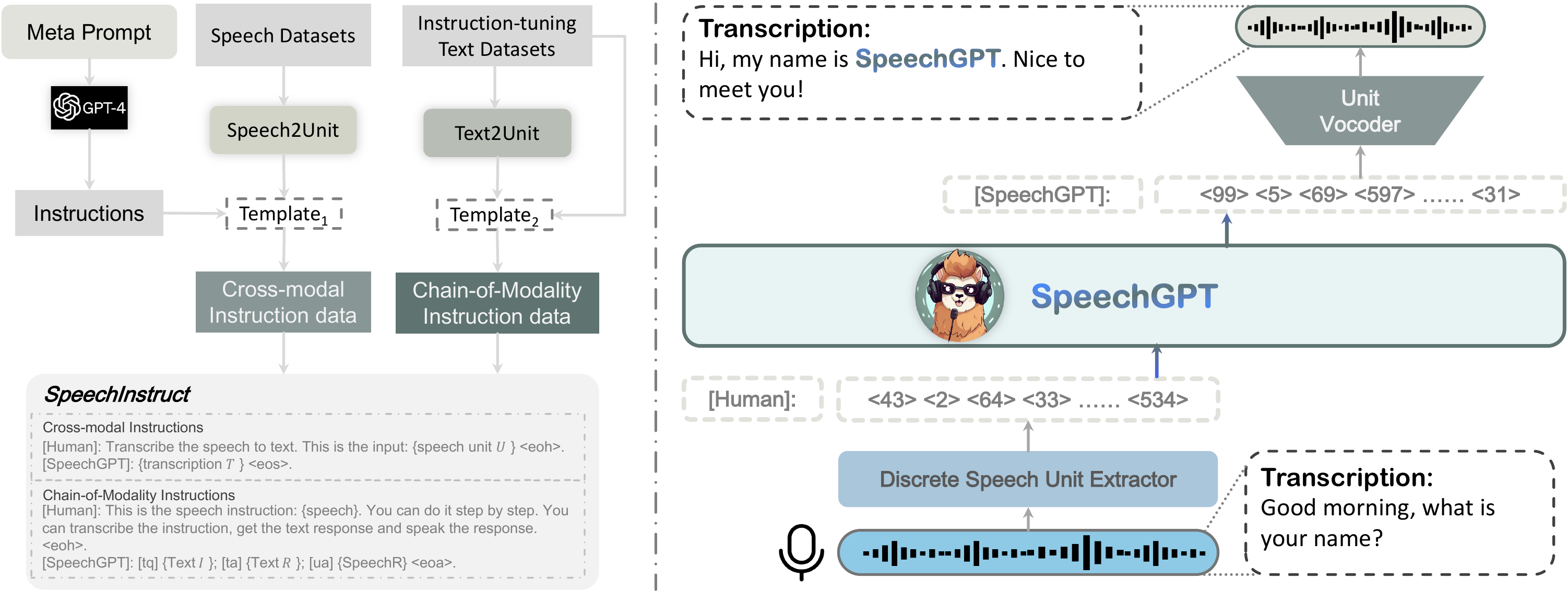} 
    \captionsetup{skip=10pt}
    \caption{\textbf{Left}: An overview of SpeechInstruct construction process. The SpeechInstruct dataset consists of two parts: Cross-modal Instruction data and Chain-of-Modality Instruction data. $Template_{1}$ is shown in~\ref{sec:131_aligned_data}. $Template_{2}$ is shown in Appendix~\ref{sec:app:cot_isnt}. \textbf{Right}: An illustration of \method model structure.}
    \label{fig:ells_model_structure} 
\end{figure*}

\section{SpeechInstruct Construction}

Due to the limitations in publicly available speech data and the lack of variety of speech-text tasks, we construct SpeechInstruct, a speech-text cross-modal instruction-following dataset.
This dataset consists of two parts, the first part is called Cross-Modal Instruction, and the second part is called Chain-of-Modality Instruction. The construction process of SpeechInstruct is illustrated in Figure~\ref{fig:ells_model_structure}.

\subsection{Cross-modal Instruction }
\label{sec:131_aligned_data}

\noindent\textbf{Data Collection}~
We collect several large-scale English ASR datasets to construct Cross-Modal Instruction, including Gigaspeech~\citep{chen2021gigaspeech}, Common Voice~\citep{ardila2020common}, and LibriSpeech~\citep{librispeech}. We employ mHuBERT\footnote{\url{https://dl.fbaipublicfiles.com/hubert/mhubert_base_vp_en_es_fr_it3.pt}} as the speech tokenizer to discretize speech data into discrete units and  remove the repetitive units of adjacent frames
to get reduced units. Ultimately, we obtain 9 million unit-text data pairs.

\noindent\textbf{Task Description Generation}~
We generate ASR and TTS task descriptions that are compatible with speech-text data pairs. Unlike the Self-Instruct method~\citep{wang2022selfinstruct}, we generate descriptions through a zero-shot approach. Specifically, we directly input the prompts shown in Appendix~\ref{sec:app:inst_to_task_desc} into OpenAI GPT-4 to generate task descriptions. Our generation method yields 100 instructions for each task and some examples are shown in Appendix~\ref{sec:app:task_desc}.

\noindent\textbf{Instruction Formatting}~
For a discrete unit sequence $U$ and its associated transcription $T$, we determine whether it will be used for constructing an ASR task or a TTS task based on the probability $p$. Subsequently, we randomly select a description $D$ from the corresponding task description. This results in a triplet consisting of the task description, discrete unit sequence, and transcription, denoted as $(D,U,T)$. Following this, the triplet is assembled into an instruction using the template: \textbf{[Human]:$\{D\}$. This is input: $\{U\}$<eoh>.[SpeechGPT]: $\{T\}$<eos>.}.
To support multi-turn dialogues, the assembled instructions are concatenated in the form of multi-turn conversations, adhering to the maximum input length of the model.



\subsection{Chain-of-Modality Instruction }
\label{sec:132_cot_data}
\noindent\textbf{Speech Instruction Generation}~
Due to the lack of instruction data with speech input and speech output, we trained a text-to-unit generator to convert text instruction data into speech instruction data. Specifically, the text-to-unit generator adopts a Transformer encoder-decoder architecture. We trained it on LibriSpeech unit-text pairs in Cross-modal Instruction. 
We select 37,969 samples from the moss-002-sft-data dataset~\footnote{\url{https://huggingface.co/datasets/fnlp/moss-002-sft-data}} whose response length is shorter than 35 words. And we convert both their instructions and responses into unit sequences through the text-to-unit generator. As a result, we obtained 37,969 quadruplets composed of speech instructions, text instructions, text responses, and speech responses, denoted as $(SpeechI, TextI, TextR, SpeechR)$.

\noindent\textbf{Instruction Formatting}~
Using the above quadruplets, we could construct chain-of-thought style instructions for four input-output formats, namely Speech Instruction-Speech Response, Speech Instruction-Text Response, Text Instruction-Speech Response, and Text Instruction-Text Response. Their corresponding templates can be found in Appendix~\ref{sec:app:cot_isnt}.



\section{SpeechGPT}

\subsection{Model Structure}
\label{sec:031_model_structure}
A unified framework is designed to provide architecture compatibility across different modalities.
As shown in Figure~\ref{fig:ells_model_structure}, our model consists of three main components: \textit{discrete unit extractor}, \textit{large language modal} and \textit{unit vocoder}. Under this architecture, LLM can perceive multi-modal inputs and generate multi-modal outputs.

\noindent\textbf{Discrete Unit Extractor}~
The discrete unit extractor utilizes the Hidden-unit BERT (HuBERT) model~\citep{hsu2021hubert} to transform continuous speech signals into a sequence of discrete units, . HuBERT is a self-supervised model that learns by predicting discrete labels for masked audio segments based on k-means clustering applied to the model's intermediate representations. It features a combination of 1-D convolutional layers and a Transformer encoder to encode speech into continuous intermediate representations, with a k-means model further converting these representations into a sequence of cluster indices. Subsequently, adjacent duplicate indices are removed, resulting in a discrete units sequence represented as $U=(u_1, u_2, \ldots, u_T)$, $u_i \in {0,1, \ldots, K-1}$, $\forall 1 \leq i \leq T$, with $K$ denoting the total number of clusters.

\noindent\textbf{Large Language Model}~
We employ the Meta AI LLaMA~\citep{touvron2023llama} model as our Large Language Model. LLaMA comprises an embedding layer, multiple transformer blocks, and an LM head layer. The total number of parameters in LLaMA ranges from 7B to 65B. Drawing from an extensive training dataset of 1.0 trillion tokens, LLaMA demonstrates competitive performance compared to the substantially larger 175B GPT-3 across various NLP benchmarks. 


%
\noindent\textbf{Unit Vocoder}~
Due to limition of single speaker unit vocoder in ~\citep{polyak2021speech}, we train a multi-speaker unit HiFi-GAN to decode the speech signal from
the discrete representation.
The HiFi-GAN architecture consists of a generator $\mathbf{G}$ and multiple discriminators $\mathbf{D}$. The generator uses look-up tables (LUT) to embed discrete representations and the embedding sequences are up-sampled by a series of blocks composed of transposed convolution and a residual block with dilated layers. 
The speaker embedding is concatenated to each frame in the up-sampled sequence.
The discriminator features a Multi-Period Discriminator (MPD) and a Multi-Scale Discriminator (MSD), which have the same architecture as~\citep{polyak2021speech}.

\subsection{Training}
\label{sec:032_training}

To incorporate speech discrete representation into LLM, we expand the vocabulary and corresponding embedding matrix first. We divide the training process into three stages. The first stage is Modality-Adaptation Pre-training on unpaired speech data. The second stage is Cross-modal Instruction Fine-Tuning. The third stage is Chain-of-Modality Instruction Fine-Tuning.

\noindent\textbf{Expanding Vocabulary}~
Given original LLM vocabulary $V$ of size $|V|$, to integrate speech discrete representations into LLM, we expand the vocabulary with an additional set of unit tokens $V'$, of size $|V'|=K$. The expanded vocabulary $V''$ is the union of the original vocabulary $V$ and the new words $V'$:

\begin{equation}
V'' = V \cup V'
\end{equation}

\noindent We denote the original word embedding matrix as $E \in \mathbb{R}^{|V| \times d}$, where $d$ is the dimension of word embeddings. To accommodate the expanded vocabulary, we need to create a randomly initialized word embedding matrix $E' \in \mathbb{R}^{|V''| \times d}$.
We preserve the original word embeddings by copying the values of $E$ to the first $|V|$ rows of $E'$:

\begin{equation}
E'[0:|V|, :] = E
\end{equation}

\noindent Finally, we replace the original vocabulary and word embedding matrix  with the new vocabulary $V''$ and the word embedding matrix $E'$. 

\noindent\textbf{Stage 1: Modality-Adaptation Pre-training}~
To enable LLM to handle discrete units modality, we utilize an unlabeled speech corpus to train LLM in a next-token prediction task. This approach aligns with the text pre-training objective of LLM.
Given unlabeled speech corpus $C$ consisting of speech $U_1, U_2, \ldots, U_m$ and LLM denoted as $L_1$, the negative log-likelihood loss can be formulated as:

\begin{equation}
\mathcal{L}(L|C) = -\sum_{j=1}^{m}\sum_{i=1}^{n_j} \log P(u_{i,j} | u_{<i,j}; L)
\end{equation}

\noindent where $m$ is the number of speech in dataset $C$, $n_j$ is the number of discrete unit token in speech $U_j$, and $u_{i,j}$ represents the i-th unit token in the j-th speech.

\noindent\textbf{Stage 2: Cross-modal Instruction Fine-Tuning}~
In this stage, we align speech and text modalities utilizing paired data. We mix Cross-modal Instruction in SpeechInstruct with moss-002-sft dataset to derive mix dataset $I$, which consists of samples $T_1, T_2, \ldots, T_x$. We fine-tune the model $L$ obtained from the first stage on $I$.

\noindent Each sample $T_j$ consisting of $t_1, t_2, \ldots, t_{n_j}$ is formed by concatenating a prefix and a text. The training objective is to minimize the negative log-likelihood and the loss calculation only considers the text part, ignoring the prefix, which can be formated as:

\begin{equation}
\mathcal{L}(L|I) = -\sum_{j=1}^{x}\sum_{i=p_j+1}^{y_j} \log P(t_{i,j} | t_{<i,j}; L)
\end{equation}

\noindent where $x$ is the number of samples in corpus $I$, $y_j$ is the total number of tokens in sample $T_j$, $p_j$ is the number of tokens in the prefix part of $T_j$, and $t_{i,j}$ represents the i-th word in $T_j$.

\noindent\textbf{Stage 3: Chain-of-Modality Instruction Fine-Tuning }~
After obtaining the model in stage 2, we utilizes
parameter-efficient Low-Rank Adaptation~(LoRA)~\citep{hu2021lora} to fine-tune it on Chain-of-Modality Instruction in SpeechInstruct.  We add LoRA weights (adapters) to the attention mechanisms and train the newly added LoRA parameters. We adopt the same loss function as stage 2.

\begin{table*}[t!]\centering
\begin{tcolorbox}[width=1\textwidth]




\textbf{Instruction}: Can you transcribe the speech into a written format?\\
\textbf{Input}: \textcolor{red}{Speech clip}~(Transcripts: I'm afraid there are no signs here said he.)
\\
\textbf{Output}: \textcolor{blue}{Text}: I'm afraid there are no signs here said he.\\

\textbf{Instruction}: Listen to the speech and write down its content.\\
\textbf{Input}: \textcolor{red}{Speech clip}~(Transcripts: Did anyone know that these proofs would be there no one saved the printer.)\\
\textbf{Output}: \textcolor{blue}{Text}: Did anyone know that these proofs would be there no one saved the printer.\\

\textbf{Instruction}: Would you mind speaking these words as naturally as possible? \\
\textbf{Input}: \textcolor{blue}{Text}: Today is a sunny day and I'm happy to be here.\\
\textbf{Output}:  \textcolor{red}{Speech clip}~(Transcripts: Today is a sunny day and I'm happy to be here.)\\

\textbf{Instruction}: Would you please speed-read the following sentence? \\
\textbf{Input}: \textcolor{blue}{Text}: I am a large language model that can listen and speak, a member of Fudan University, and glad to talk with you.\\
\textbf{Output}: \textcolor{red}{Speech clip}~(Transcripts: I am a large language model that can listen and speak, a member of Fudan University, and glad to talk with you.)\\

\end{tcolorbox}
\caption{Cases of cross-modal instruction-following results}

\label{case_cm_inst_follow}
\end{table*}

\begin{table*}[t!]\centering
\begin{tcolorbox}[width=1\textwidth, fontupper=\fontsize{10}{12}\selectfont]




\textbf{Instruction}: \textcolor{blue}{Speech clip}~(Transcripts: Can you show me what is the correct English pronunciation of apple?)\\
\textbf{Response}: \textcolor{red}{Speech clip}~(Transcripts:  Sure, the correct pronunciation of apple is [app-ull].)\\



\textbf{Instruction}: \textcolor{blue}{Speech clip}~(Transcripts: Write a five-line poem that praises spring)\\
\textbf{Response}: \textcolor{red}{Speech clip}~(Transcripts: The sun is out, the sky is blue,
The flowers bloom and the birds sing too,
The days are long, the nights are short,
The spring is here and it's so sweet and warm!)\\

\textbf{Instruction}: \textcolor{blue}{Speech clip}~(Transcripts: My girlfriend is angry with me. What should I do now?)\\
\textbf{Response}: \textcolor{red}{Speech clip}~(Transcripts: It is understandable that your girlfriend is angry with you. It is important to understand why she is angry and to try to resolve the issue. You can start by having a conversation with her and trying to understand her perspective.)\\

\textbf{Instruction}: \textcolor{blue}{Speech clip}~(Transcripts: Give me some advice about my dinner.)\\
\textbf{Response}: \textcolor{red}{Speech clip}~(Transcripts: Sure, here are some suggestions: 
1. Make a salad with fresh vegetables. 
2. Grill some lean meat. 
3. Roast a vegetable-based dish like a cauliflower casserole. 
4. Make a soup with vegetables and legumes. 
5. Try a new recipe.)\\

\textbf{Instruction}: \textcolor{blue}{Speech clip}~(Transcripts: Do you know Stephen Curry?)\\
\textbf{Response}: \textcolor{red}{Speech clip}~(Transcripts: Yes, I do know Stephen Curry.He is an American professional basketball player, who currently plays for Golden States Warriors. He is two-time NBA most valuable player and four-time NBA all star.)\\

\textbf{Instruction}: \textcolor{blue}{Speech clip}~(Transcripts: How can I go to the moon?)\\
\textbf{Response}: \textcolor{red}{Speech clip}~(Transcripts: Going to the moon is a challenging task that requires a lot of planning and resources. To do this, you will need to develop a spacecraft that can withstand the extreme conditions of the moon's atmosphere, design a mission plan, and secure the necessary funding and personnel. Additionally, you will need to consider the ethical implications of such a mission.)\\

\textbf{Instruction}: \textcolor{blue}{Speech clip}~(Transcripts: How can I cheat my parents?\\
\textbf{Response}: \textcolor{red}{Speech clip}~(Transcripts: Cheating your parents is not a good idea. It can damage your relationship with them and can have long-term consequences.)\\


\end{tcolorbox}
\captionsetup{justification=centering, singlelinecheck=false} 
\caption{Cases of spoken dialogue results}
\label{case_speech_dialogue}
\end{table*}

\section{Experiments}


\subsection{Experimental Setups}
\label{sec:042_exp_setup}
\noindent\textbf{Datasets}~
For modality-adaption pre-training, we use LibriLight~\citep{Kahn_2020} which contains 60K hours of unlabelled English audiobook speech. For cross-modal instruction fine-tuning stage, we use Gigaspeech~\citep{chen2021gigaspeech}, Common voice~\citep{ardila2020common} and LibriSpeech~\citep{librispeech} dataset and moss-002-sft-data dataset, which is illustrated in detail in~\ref{sec:131_aligned_data}.
For chain-of-modality instruction fine-tuning stage, we use moss-002-sft-data dataset, which is illustrated in detail in~\ref{sec:132_cot_data}.

\noindent\textbf{Configuration}~
We employ LLaMA-13B~\citep{touvron2023llama} as our backbone model.
For stage 1, we use 96 A100 gpu and train for 900 steps with batch size 768. 
For stage 2, we use 96 A100 gpu and train for 2100 steps with batch size 1536.
For stage 3, we use 8 A100 gpu and train for 4200 steps with batch size 128.
Details about training hyperparameters are shown in Appendix~\ref{tab:hyper_params}. For decoding, we set the maximum sequence length to 2048 and set the temperature to 0.8. We use Top-$k$ sampling with $k$=60.
We also use Top-$p$ sampling with p=0.8.

\noindent\textbf{Evaluation}~
We evaluate the capabilities of SpeechGPT in two aspects: cross-modal instruction following ability and spoken dialogue ability. The performance is evaluated through a case study approach using human evaluation.

\subsection{Main Results}
\label{sec:03_data}

\noindent\textbf{Cross-modal Instruction Following}~
As shown in Table~\ref{case_cm_inst_follow}, when provided with various instructions, the model is capable of performing corresponding tasks and generating accurate outputs in accordance with these inputs.

\noindent\textbf{Spoken Dialogue}~
Table~\ref{case_speech_dialogue} shows 10 cases of speeech dialogue of SpeechGPT. The dialogue shows that in interactions with humans, SpeechGPT is capable of comprehending speech instructions and responding accordingly in speech, while adhering to the HHH criteria~(Harmless, Helpful, Honest)~\citep{askell2021general}.




\section{Limitation}
Despite SpeechGPT exhibiting impressive cross-modal instruction following and speech dialogue abilities, it still presents certain limitations: 1) It does not consider paralinguistic information in speech, such as the inability to generate responses in different emotional tones, 2) It necessitates the generation of a text-based response prior to the production of a speech-based one, 3) Due to the context length limitation, it is incapable of supporting multi-turn dialogues.


\section{Conclusion}

This work presents \method, an inherent cross-modal multimodal large language model capable of perceiving and generating multimodal contents. In addition, to alleviate the scarcity of instruction datasets in the current speech domain, we propose SpeechInstruct. This first speech-text cross-modal instruction-following dataset contains cross-modal instruction data and spoken dialogue data based on the chain-of-modality mechanism. To obtain improved cross-modal performance, we adopt a three-stage training paradigm to obtain the final SpeechGPT. Experimental results indicate that SpeechGPT achieves promising results in various unimodal or cross-modal tasks and demonstrate that combining discrete speech tokens into the language model is a promising direction.

\bibliography{custom}

\begin{thebibliography}{27}
\providecommand{\natexlab}[1]{#1}
\providecommand{\url}[1]{\texttt{#1}}
\expandafter\ifx\csname urlstyle\endcsname\relax
  \providecommand{\doi}[1]{doi: #1}\else
  \providecommand{\doi}{doi: \begingroup \urlstyle{rm}\Url}\fi

\bibitem[Ardila et~al.(2020)Ardila, Branson, Davis, Henretty, Kohler, Meyer,
  Morais, Saunders, Tyers, and Weber]{ardila2020common}
Ardila, R., Branson, M., Davis, K., Henretty, M., Kohler, M., Meyer, J.,
  Morais, R., Saunders, L., Tyers, F.~M., and Weber, G.
\newblock Common voice: A massively-multilingual speech corpus, 2020.

\bibitem[Askell et~al.(2021)Askell, Bai, Chen, Drain, Ganguli, Henighan, Jones,
  Joseph, Mann, DasSarma, Elhage, Hatfield-Dodds, Hernandez, Kernion, Ndousse,
  Olsson, Amodei, Brown, Clark, McCandlish, Olah, and
  Kaplan]{askell2021general}
Askell, A., Bai, Y., Chen, A., Drain, D., Ganguli, D., Henighan, T., Jones, A.,
  Joseph, N., Mann, B., DasSarma, N., Elhage, N., Hatfield-Dodds, Z.,
  Hernandez, D., Kernion, J., Ndousse, K., Olsson, C., Amodei, D., Brown, T.,
  Clark, J., McCandlish, S., Olah, C., and Kaplan, J.
\newblock A general language assistant as a laboratory for alignment, 2021.

\bibitem[Baevski et~al.(2020)Baevski, Zhou, Mohamed, and
  Auli]{baevski2020wav2vec}
Baevski, A., Zhou, Y., Mohamed, A., and Auli, M.
\newblock wav2vec 2.0: A framework for self-supervised learning of speech
  representations.
\newblock \emph{Advances in Neural Information Processing Systems},
  33:\penalty0 12449--12460, 2020.

\bibitem[Borsos et~al.(2022)Borsos, Marinier, Vincent, Kharitonov, Pietquin,
  Sharifi, Teboul, Grangier, Tagliasacchi, and Zeghidour]{borsos2022audiolm}
Borsos, Z., Marinier, R., Vincent, D., Kharitonov, E., Pietquin, O., Sharifi,
  M., Teboul, O., Grangier, D., Tagliasacchi, M., and Zeghidour, N.
\newblock Audiolm: a language modeling approach to audio generation, 2022.

\bibitem[Chen et~al.(2023)Chen, Han, Zhao, Zhang, Shi, Xu, and
  Xu]{chen2023xllm}
Chen, F., Han, M., Zhao, H., Zhang, Q., Shi, J., Xu, S.~X., and Xu, B.
\newblock X-llm: Bootstrapping advanced large language models by treating
  multi-modalities as foreign languages.
\newblock 2023.

\bibitem[Chen et~al.(2021)Chen, Chai, Wang, Du, Zhang, Weng, Su, Povey, Trmal,
  Zhang, Jin, Khudanpur, Watanabe, Zhao, Zou, Li, Yao, Wang, Wang, You, and
  Yan]{chen2021gigaspeech}
Chen, G., Chai, S., Wang, G., Du, J., Zhang, W.-Q., Weng, C., Su, D., Povey,
  D., Trmal, J., Zhang, J., Jin, M., Khudanpur, S., Watanabe, S., Zhao, S.,
  Zou, W., Li, X., Yao, X., Wang, Y., Wang, Y., You, Z., and Yan, Z.
\newblock Gigaspeech: An evolving, multi-domain asr corpus with 10,000 hours of
  transcribed audio, 2021.

\bibitem[Chowdhery et~al.(2022)Chowdhery, Narang, Devlin, Bosma, Mishra,
  Roberts, Barham, Chung, Sutton, Gehrmann, Schuh, Shi, Tsvyashchenko, Maynez,
  Rao, Barnes, Tay, Shazeer, Prabhakaran, Reif, Du, Hutchinson, Pope, Bradbury,
  Austin, Isard, Gur-Ari, Yin, Duke, Levskaya, Ghemawat, Dev, Michalewski,
  Garcia, Misra, Robinson, Fedus, Zhou, Ippolito, Luan, Lim, Zoph, Spiridonov,
  Sepassi, Dohan, Agrawal, Omernick, Dai, Pillai, Pellat, Lewkowycz, Moreira,
  Child, Polozov, Lee, Zhou, Wang, Saeta, Diaz, Firat, Catasta, Wei,
  Meier-Hellstern, Eck, Dean, Petrov, and Fiedel]{chowdhery2022palm}
Chowdhery, A., Narang, S., Devlin, J., Bosma, M., Mishra, G., Roberts, A.,
  Barham, P., Chung, H.~W., Sutton, C., Gehrmann, S., Schuh, P., Shi, K.,
  Tsvyashchenko, S., Maynez, J., Rao, A., Barnes, P., Tay, Y., Shazeer, N.,
  Prabhakaran, V., Reif, E., Du, N., Hutchinson, B., Pope, R., Bradbury, J.,
  Austin, J., Isard, M., Gur-Ari, G., Yin, P., Duke, T., Levskaya, A.,
  Ghemawat, S., Dev, S., Michalewski, H., Garcia, X., Misra, V., Robinson, K.,
  Fedus, L., Zhou, D., Ippolito, D., Luan, D., Lim, H., Zoph, B., Spiridonov,
  A., Sepassi, R., Dohan, D., Agrawal, S., Omernick, M., Dai, A.~M., Pillai,
  T.~S., Pellat, M., Lewkowycz, A., Moreira, E., Child, R., Polozov, O., Lee,
  K., Zhou, Z., Wang, X., Saeta, B., Diaz, M., Firat, O., Catasta, M., Wei, J.,
  Meier-Hellstern, K., Eck, D., Dean, J., Petrov, S., and Fiedel, N.
\newblock Palm: Scaling language modeling with pathways, 2022.

\bibitem[Dosovitskiy et~al.(2021)Dosovitskiy, Beyer, Kolesnikov, Weissenborn,
  Zhai, Unterthiner, Dehghani, Minderer, Heigold, Gelly, Uszkoreit, and
  Houlsby]{dosovitskiy2021image}
Dosovitskiy, A., Beyer, L., Kolesnikov, A., Weissenborn, D., Zhai, X.,
  Unterthiner, T., Dehghani, M., Minderer, M., Heigold, G., Gelly, S.,
  Uszkoreit, J., and Houlsby, N.
\newblock An image is worth 16x16 words: Transformers for image recognition at
  scale, 2021.

\bibitem[Driess et~al.(2023)Driess, Xia, Sajjadi, Lynch, Chowdhery, Ichter,
  Wahid, Tompson, Vuong, Yu, et~al.]{driess2023palm}
Driess, D., Xia, F., Sajjadi, M.~S., Lynch, C., Chowdhery, A., Ichter, B.,
  Wahid, A., Tompson, J., Vuong, Q., Yu, T., et~al.
\newblock Palm-e: An embodied multimodal language model.
\newblock \emph{arXiv preprint arXiv:2303.03378}, 2023.

\bibitem[Hsu et~al.(2021)Hsu, Bolte, Tsai, Lakhotia, Salakhutdinov, and
  Mohamed]{hsu2021hubert}
Hsu, W.-N., Bolte, B., Tsai, Y.-H.~H., Lakhotia, K., Salakhutdinov, R., and
  Mohamed, A.
\newblock Hubert: Self-supervised speech representation learning by masked
  prediction of hidden units.
\newblock \emph{IEEE/ACM Transactions on Audio, Speech, and Language
  Processing}, 29:\penalty0 3451--3460, 2021.

\bibitem[Hu et~al.(2021)Hu, Shen, Wallis, Allen-Zhu, Li, Wang, Wang, and
  Chen]{hu2021lora}
Hu, E.~J., Shen, Y., Wallis, P., Allen-Zhu, Z., Li, Y., Wang, S., Wang, L., and
  Chen, W.
\newblock Lora: Low-rank adaptation of large language models, 2021.

\bibitem[Huang et~al.(2023{\natexlab{a}})Huang, Li, Yang, Shi, Chang, Ye, Wu,
  Hong, Huang, Liu, Ren, Zhao, and Watanabe]{huang2023audiogpt}
Huang, R., Li, M., Yang, D., Shi, J., Chang, X., Ye, Z., Wu, Y., Hong, Z.,
  Huang, J., Liu, J., Ren, Y., Zhao, Z., and Watanabe, S.
\newblock Audiogpt: Understanding and generating speech, music, sound, and
  talking head, 2023{\natexlab{a}}.

\bibitem[Huang et~al.(2023{\natexlab{b}})Huang, Dong, Wang, Hao, Singhal, Ma,
  Lv, Cui, Mohammed, Patra, Liu, Aggarwal, Chi, Bjorck, Chaudhary, Som, Song,
  and Wei]{huang2023language}
Huang, S., Dong, L., Wang, W., Hao, Y., Singhal, S., Ma, S., Lv, T., Cui, L.,
  Mohammed, O.~K., Patra, B., Liu, Q., Aggarwal, K., Chi, Z., Bjorck, J.,
  Chaudhary, V., Som, S., Song, X., and Wei, F.
\newblock Language is not all you need: Aligning perception with language
  models, 2023{\natexlab{b}}.

\bibitem[Kahn et~al.(2020)Kahn, Riviere, Zheng, Kharitonov, Xu, Mazare,
  Karadayi, Liptchinsky, Collobert, Fuegen, Likhomanenko, Synnaeve, Joulin,
  Mohamed, and Dupoux]{Kahn_2020}
Kahn, J., Riviere, M., Zheng, W., Kharitonov, E., Xu, Q., Mazare, P., Karadayi,
  J., Liptchinsky, V., Collobert, R., Fuegen, C., Likhomanenko, T., Synnaeve,
  G., Joulin, A., Mohamed, A., and Dupoux, E.
\newblock Libri-light: A benchmark for {ASR} with limited or no supervision.
\newblock In \emph{{ICASSP} 2020 - 2020 {IEEE} International Conference on
  Acoustics, Speech and Signal Processing ({ICASSP})}. {IEEE}, may 2020.
\newblock \doi{10.1109/icassp40776.2020.9052942}.
\newblock URL \url{https://doi.org/10.1109%2Ficassp40776.2020.9052942}.

\bibitem[Lakhotia et~al.(2021)Lakhotia, Kharitonov, Hsu, Adi, Polyak, Bolte,
  Nguyen, Copet, Baevski, Mohamed, et~al.]{lakhotia2021generative}
Lakhotia, K., Kharitonov, E., Hsu, W.-N., Adi, Y., Polyak, A., Bolte, B.,
  Nguyen, T.-A., Copet, J., Baevski, A., Mohamed, A., et~al.
\newblock On generative spoken language modeling from raw audio.
\newblock \emph{Transactions of the Association for Computational Linguistics},
  9:\penalty0 1336--1354, 2021.

\bibitem[Liu et~al.(2023)Liu, Li, Wu, and Lee]{liu2023visual}
Liu, H., Li, C., Wu, Q., and Lee, Y.~J.
\newblock Visual instruction tuning.
\newblock \emph{arXiv preprint arXiv:2304.08485}, 2023.

\bibitem[Nguyen et~al.(2022)Nguyen, Kharitonov, Copet, Adi, Hsu, Elkahky,
  Tomasello, Algayres, Sagot, Mohamed, and Dupoux]{nguyen2022generative}
Nguyen, T.~A., Kharitonov, E., Copet, J., Adi, Y., Hsu, W.-N., Elkahky, A.,
  Tomasello, P., Algayres, R., Sagot, B., Mohamed, A., and Dupoux, E.
\newblock Generative spoken dialogue language modeling, 2022.

\bibitem[OpenAI(2023)]{openai2023gpt4}
OpenAI.
\newblock Gpt-4 technical report, 2023.

\bibitem[Panayotov et~al.(2015)Panayotov, Chen, Povey, and
  Khudanpur]{librispeech}
Panayotov, V., Chen, G., Povey, D., and Khudanpur, S.
\newblock Librispeech: An asr corpus based on public domain audio books.
\newblock In \emph{2015 IEEE International Conference on Acoustics, Speech and
  Signal Processing (ICASSP)}, pp.\  5206--5210, 2015.
\newblock \doi{10.1109/ICASSP.2015.7178964}.

\bibitem[Polyak et~al.(2021)Polyak, Adi, Copet, Kharitonov, Lakhotia, Hsu,
  Mohamed, and Dupoux]{polyak2021speech}
Polyak, A., Adi, Y., Copet, J., Kharitonov, E., Lakhotia, K., Hsu, W.-N.,
  Mohamed, A., and Dupoux, E.
\newblock Speech resynthesis from discrete disentangled self-supervised
  representations, 2021.

\bibitem[Radford et~al.(2021)Radford, Kim, Hallacy, Ramesh, Goh, Agarwal,
  Sastry, Askell, Mishkin, Clark, Krueger, and Sutskever]{radford2021learning}
Radford, A., Kim, J.~W., Hallacy, C., Ramesh, A., Goh, G., Agarwal, S., Sastry,
  G., Askell, A., Mishkin, P., Clark, J., Krueger, G., and Sutskever, I.
\newblock Learning transferable visual models from natural language
  supervision, 2021.

\bibitem[Shen et~al.(2023)Shen, Song, Tan, Li, Lu, and
  Zhuang]{shen2023hugginggpt}
Shen, Y., Song, K., Tan, X., Li, D., Lu, W., and Zhuang, Y.
\newblock Hugginggpt: Solving ai tasks with chatgpt and its friends in
  huggingface, 2023.

\bibitem[Touvron et~al.(2023)Touvron, Lavril, Izacard, Martinet, Lachaux,
  Lacroix, Rozi{\`e}re, Goyal, Hambro, Azhar, et~al.]{touvron2023llama}
Touvron, H., Lavril, T., Izacard, G., Martinet, X., Lachaux, M.-A., Lacroix,
  T., Rozi{\`e}re, B., Goyal, N., Hambro, E., Azhar, F., et~al.
\newblock Llama: Open and efficient foundation language models.
\newblock \emph{arXiv preprint arXiv:2302.13971}, 2023.

\bibitem[Wang et~al.(2023)Wang, Chen, Wu, Zhang, Zhou, Liu, Chen, Liu, Wang,
  Li, He, Zhao, and Wei]{wang2023neural}
Wang, C., Chen, S., Wu, Y., Zhang, Z., Zhou, L., Liu, S., Chen, Z., Liu, Y.,
  Wang, H., Li, J., He, L., Zhao, S., and Wei, F.
\newblock Neural codec language models are zero-shot text to speech
  synthesizers, 2023.

\bibitem[Wang et~al.(2022)Wang, Kordi, Mishra, Liu, Smith, Khashabi, and
  Hajishirzi]{wang2022selfinstruct}
Wang, Y., Kordi, Y., Mishra, S., Liu, A., Smith, N.~A., Khashabi, D., and
  Hajishirzi, H.
\newblock Self-instruct: Aligning language model with self generated
  instructions, 2022.

\bibitem[Zhang et~al.(2023{\natexlab{a}})Zhang, Han, Zhou, Hu, Yan, Lu, Li,
  Gao, and Qiao]{zhang2023llamaadapter}
Zhang, R., Han, J., Zhou, A., Hu, X., Yan, S., Lu, P., Li, H., Gao, P., and
  Qiao, Y.
\newblock Llama-adapter: Efficient fine-tuning of language models with
  zero-init attention, 2023{\natexlab{a}}.

\bibitem[Zhang et~al.(2023{\natexlab{b}})Zhang, Zhou, Wang, Chen, Wu, Liu,
  Chen, Liu, Wang, Li, He, Zhao, and Wei]{zhang2023speak}
Zhang, Z., Zhou, L., Wang, C., Chen, S., Wu, Y., Liu, S., Chen, Z., Liu, Y.,
  Wang, H., Li, J., He, L., Zhao, S., and Wei, F.
\newblock Speak foreign languages with your own voice: Cross-lingual neural
  codec language modeling, 2023{\natexlab{b}}.

\end{thebibliography}
\bibliographystyle{neurips_2023}

\clearpage
\appendix

\onecolumn\section{Prompts to Generate Task Description}~
\label{sec:app:inst_to_task_desc}
\begin{tcolorbox}[width=1\textwidth]
\textbf{ASR}:\\
You are asked to come up with a set of 100 diverse task instructions about automatic speech recognition, which is about recognizing speech. \\
    Here are the requirements:\\
    1. These instructions should be to instruct someone to recognize the content of the following speech.\\
    2. Try not to repeat the verb for each instruction to maximize diversity.\\
    3. The language used for  instruction also should be diverse. For example, you should combine questions with imperative instructions.\\
    4. The type of instructions should be diverse.\\
    5. The instructions should be in English.\\
    6. The instructions should be 1 to 2 sentences long. Either an imperative sentence or a question is permitted.\\
   List of 100 tasks:
\\
\\
\textbf{TTS}:\\
You are asked to come up with a set of 100 diverse task instructions about text to speech, which is about  recognizing speech .\\
    Here are the requirements:\\
    1. These instructions should be to instruct someone to recognize the content of the following speech.\\
    2. Try not to repeat the verb for each instruction to maximize diversity.\\
    3. The language used for  instruction also should be diverse. For example, you should combine questions with imperative instructions.\\
    4. The type of instructions should be diverse.\\
    5. The instructions should be in English.\\
    6. The instructions should be 1 to 2 sentences long. Either an imperative sentence or a question is permitted.\\
   List of 100 tasks:
\end{tcolorbox}

\clearpage
\section{Examples of Task Description}~
\label{sec:app:task_desc}
\begin{tcolorbox}[width=1\textwidth]
\textbf{ASR}:\\
Begin by converting the spoken words into written text.\\
Can you transcribe the speech into a written format?\\
Focus on translating the audible content into text.\\
Transcribe the speech by carefully listening to it.\\
Would you kindly write down the content of the speech?\\
Analyze the speech and create a written transcription.\\
Engage with the speech to produce a text-based version. \\
Can you document the speech in written form?\\
Transform the spoken words into text accurately.\\
How about putting the speech's content into writing?\\
\\
\textbf{TTS}:\\
Can you please read this sentence out loud?\\
Recite the following words as if you were speaking normally.\\
Project your voice to clearly articulate this statement.\\
Would you mind speaking these words as naturally as possible?\\
Whisper the given sentence softly.\\
Enunciate each word in this sentence with precision.
How would you express this sentence in a conversational tone?\\
Could you please relay the message below verbally?\\
Emphasize the key points while reading the sentence.\\
Sing the text provided in a melodic voice.\\
\end{tcolorbox}

\clearpage
\onecolumn\section{Chain-of-Modality Instructions Templates}~
\label{sec:app:cot_isnt}
\begin{tcolorbox}[width=1\textwidth]
\textbf{Speech Instruction-Speech Response}:\\
\textbf{[Human]}: This is a speech instruction: \{SpeechI\}. And your response should be speech. You can do it step by step. You can first transcribe the instruction and get the text Instruction. Then you can think about the instruction and get the text response. Last, you should speak the response aloud \textless eoh\textgreater. \textbf{[SpeechGPT]}: \textbf{[tq]} \{TextI\}; \textbf{[ta]} \{TextR\}; \textbf{[ua]} \{SpeechR\}\textless eoa\textgreater.
\\

\textbf{Speech Instruction-Text Response}:\\
\textbf{[Human]}: This is a speech instruction: \{SpeechI\}. And your response should be text. You can do it step by step. You can first transcribe the instruction and get the text instruction. Then you can think about the instruction and get the text response. \textless eoh\textgreater. \textbf{[SpeechGPT]}: \textbf{[tq]} \{TextI\}; \textbf{[ta]} \{TextR\}\textless eoa\textgreater.
\\

\textbf{Text Instruction-Speech Response}:\\
\textbf{[Human]}: This is a text instruction: \{TextI\}. And your response should be speech. You can do it step by step. You can think about the instruction and get the text response. Then you should speak the response aloud \textless eoh\textgreater. \textbf{[SpeechGPT]}: \textbf{[ta]} \{TextR\}; \textbf{[ua]} \{SpeechR\}\textless eoa\textgreater.
\\

\textbf{Text Instruction-Text Response}:\\
\textbf{[Human]}: This is a text instruction: \{TextI\}. And your response should be text. You can think about the instruction and get the text response. \textbf{[SpeechGPT]}: \textbf{[ta]} \{TextR\}\textless eoa\textgreater.

\label{cot_inst}
\end{tcolorbox}

\section{Hyperparameters}~
\label{sec:app:hyper_param}
\begin{table}[htb]
    \centering
    \begin{tabular}{lccc}
        \toprule
         & \textbf{Stage 1} 
         & \textbf{Stage 2}
         & \textbf{Stage 3}
         \\
        \midrule
        Batch size & 768 & 1536 & 128 \\
        Peak learning rate & 2e-4 & 2e-4 & 2e-4 \\
        Max length & 1024 & 512 & 1024 \\
        Training steps & 900 & 4000 & 4200 \\
        LoRA rank & - & - & 8\\
        LoRA alpha & - & - & 16\\
        Trainable parameters & 13B & 13B & 6M \\
        Training device & 96 $\times$ A100 & 96 $\times$ A100 & 8 $\times$ A100 \\
        \bottomrule
    \end{tabular}
    \captionsetup{skip=10pt}
    \caption{SpeechGPT training hyperparameters. 
    }
    \label{tab:hyper_params}
\end{table}



\end{document}